\documentclass[10pt,twocolumn,letterpaper]{article}
\usepackage{graphicx} 
\usepackage{authblk}

\usepackage[pagenumbers]{cvpr} 

\usepackage{amsmath}
\usepackage{placeins}

\usepackage{newtxtext,newtxmath}
\usepackage{booktabs}
\usepackage[pagebackref,breaklinks,colorlinks]{hyperref}
\usepackage[capitalize,nameinlink,noabbrev]{cleveref}

\usepackage{adjustbox}  
\usepackage{makecell}
\usepackage{multirow}
\usepackage{booktabs}
\usepackage{pifont}
\usepackage{flushend}
\usepackage{titletoc}

\newcommand{\cmark}{\ding{51}}
\newcommand{\xmark}{\ding{55}}
\newcommand{\rot}[1]{\rotatebox[origin=c]{90}{#1}}
\definecolor{cutePink}{RGB}{242, 69, 153}  

\newcommand{\PAR}[1]{\vspace{-0.1eM}\vskip4pt \noindent{\bf #1}}

\definecolor{cutePink}{RGB}{255, 182, 193}

\title{ScenarioControl:\\ Vision-Language Controllable Vectorized Latent Scenario Generation}

\author{
Lili Gao$^{1,*}$, 
Yanbo Xu$^{2,*}$, 
William Koch$^{2,*}$, 
Samuele Ruffino$^{1}$, 
Luke Rowe$^{3}$, 
Behdad Chalaki$^{1}$, \quad
Dmitriy Rivkin$^{1}$,
Julian Ost$^{1,2}$,
Roger Girgis$^{1,3}$, 
Mario Bijelic$^{1,2}$ 
and Felix Heide$^{1,2}$ \\
$^1$Torc Robotics,\ \ 
$^2$Princeton University,\ \ 
$^3$Mila 
}

\begin{document}
\maketitle

\begin{abstract}
We introduce ScenarioControl, the first vision-language control mechanism for learned driving scenario generation. Given a text prompt or an input image, ScenarioControl synthesizes diverse, realistic 3D scenario rollouts -- including map, 3D boxes of reactive actors over time, pedestrians, driving infrastructure, and ego camera observations. The method generates scenes in a vectorized latent space that represents road structure and dynamic agents jointly. To connect multimodal control with sparse vectorized scene elements, we propose a cross-global control mechanism that integrates cross-attention with a lightweight global-context branch, enabling fine-grained control over road layout and traffic conditions while preserving realism. The method produces temporally consistent scenario rollouts from the perspectives different actors in the scene, supporting long-horizon continuation of driving scenarios. To facilitate training and evaluation, we release a dataset with text annotations aligned to vectorized map structures. Extensive experiments validate that the control adherence and fidelity of ScenarioControl compare favorable to all tested methods across all experiments. Project webpage: \href{https://light.princeton.edu/ScenarioControl}{\color{cutePink}{https://light.princeton.edu/ScenarioControl}}

\end{abstract}   
\renewcommand*{\thefootnote}{\fnsymbol{footnote}} \footnotetext{$^*$Indicates equal contribution.}
\section{Introduction}
\label{sec:intro}

Large-scale datasets have been indispensable for the progress of autonomous driving~\cite{hu2023_uniad}, providing multi-modal labeled data across regions and conditions \cite{nuscenes2019,Geiger2013IJRRkitti,Sun_2020_CVPR,Sun_2020_CVPR_waymo_perception,Karnchanachari2024TowardsLPnuplan}. 
However, real-world logs alone are insufficient to capture rare but safety-critical events, such as wrong-way drivers. Evaluating these edge cases is essential for safe and reliable systems, yet relying solely on collected data is highly sample-inefficient~\cite{Scanlon2021WaymoSD}.

Simulators bridge this gap by enabling safe, repeatable, and scalable experimentation. Traditional rule-based simulators such as CARLA~\cite{dosovitskiy2017carlaopenurbandriving}, SMARTS~\cite{zhou2020smartsscalablemultiagentreinforcement}, and MetaDrive~\cite{li2022metadrivecomposingdiversedriving} provide controllable virtual worlds for perception and planning research, yet their handcrafted world design limits realism and diversity, particularly for rare events \cite{simulationreview_li2024,datadrivensimulation_chen2024}. 
Recent generative approaches, such as Driving Diffusion \cite{li2023drivingdiffusion} and Panacea \cite{wen2024panacea}, introduce controllability by conditioning on structured ``control layouts'' derived from existing scenarios which steer generation by projecting logged scenes into intermediate control representations \cite{li2023drivingdiffusion,wen2024panacea,lin2025drivegen,gao2025magicdrive-v2,nvidia2025cosmosdrivedreams,wang2023drivedreamer,hu2023gaia1generativeworldmodel,russell2025gaia2controllablemultiviewgenerative}. 
Behavior- and interaction-level methods control agent dynamics to create challenging closed-loop interactions, but commonly assume a given road graph and initial scene context \cite{rowe2024ctrlsim,zhou2024behaviorgpt,gulino2023waymaxaccelerateddatadrivensimulator,feng2023trafficgen}. 
Even for a given initial scenario description, these methods often struggle to generate long-tail events as the datasets used to train them contain few of such examples.
To the best of our knowledge, no existing method allows for vision-language-controlled scenario generation.

In parallel, diffusion-based and data-driven simulation work has made substantial progress on generating realistic and diverse driving scenes from data.
These methods treat scenario synthesis as a generative modeling problem over vectorized~\cite{rowe2025scenariodreamervectorizedlatent} or rasterized representations~\cite{chitta2024sledgesynthesizingdrivingenvironments,pronovost2023scenario,scenecontrol2024,datadrivensimulation_chen2024}, either placing actors on an existing road graph~\cite{scenecontrol2024,pronovost2023scenario} or generating both the road graph and actor positions jointly~\cite{rowe2025scenariodreamervectorizedlatent,chitta2024sledgesynthesizingdrivingenvironments}. 
Generative models have been shown to produce high-fidelity structured outputs -- road topology and traffic participants that are directly consumable by downstream components such as motion planning, sensor simulation, end-to-end driving, and multimodal future generation \cite{gulino2023waymaxaccelerateddatadrivensimulator,zhou2024behaviorgpt,liao2024diffusiondrive,Wang2025DiffAD,Guan2025WorldME,umgen2025wu}. 
However, a crucial gap remains: while realism and diversity are consistently improving, the controllability at the scenario-level of the layouts themselves is still limited. Generation is either entirely uncontrolled - by sampling from the latent space - or with control signals taken from existing scenarios, and extrapolation does not expose interpretable control knobs \cite{chitta2024sledgesynthesizingdrivingenvironments,rowe2025scenariodreamervectorizedlatent}. We propose ScenarioControl to bridge this gap. ScenarioControl enables vision-language control of diffusion-based generation, unlocking controllable synthesis of structured driving scenarios and camera sensor simulation for the ego, and any other actor in the scene. Unlike other methods, it does so without relying on given logged scene configurations or auto-labeled control layouts. The resulting scenarios are plug-and-play with established simulators and autonomy stacks. 

To this end, we introduce a novel cross-global control mechanism that conditions sparse vectorized scene tokens on dense features, from either a text prompt or a dashcam ego image, via two complementary branches: a cross-attention branch for fine-grained control and a lightweight global-context branch for capturing high-level scene intent. Conditioning on natural language and visual cues enables goal-directed scenario synthesis that both reflects real-world context and targets specific long-tail regimes. By steering road geometry, agent placement, and traffic conditions with a text prompt or a single ego dashcam-style image, generation moves beyond unconstrained sampling while preserving realistic structure and dynamics, crucial for synthesizing safety-critical cases and building targeted evaluation/training sets. In addition to enabling prompt- and image-conditioned scene generation, ScenarioControl also supports scene outpainting and long-horizon video continuation, maintaining temporal and visual consistency. We confirm ScenarioControl's fidelity and controllability with quantitative experiments, while qualitative results demonstrate adherence to conditioning and diverse long-horizon rollouts.

\noindent Our contributions are summarized as follows:
\begin{itemize}
  \item We propose a vision–language conditioned vectorized latent diffusion model that generates full vectorized driving scenes, including lane topology, agent placement, and traffic signals conditioned on text prompts or dash-cam-style ego images.
\item We introduce a new conditioning mechanism that fuses sparse, vectorized road layouts with dense prompt and image representations, enabling fine-grained control over generation.
   \item We evaluate controllability, diversity, and fidelity of the generated scenarios for arbitrary actors in the scene, confirming that the proposed method performs favorably compared to existing methods while providing fine-grained vision-language control.
\end{itemize}

\section{Related Work}
\label{sec:rel_work}
\begin{figure*}[t]
\vspace{-5pt}
\centering
\includegraphics[width=1.0\linewidth]{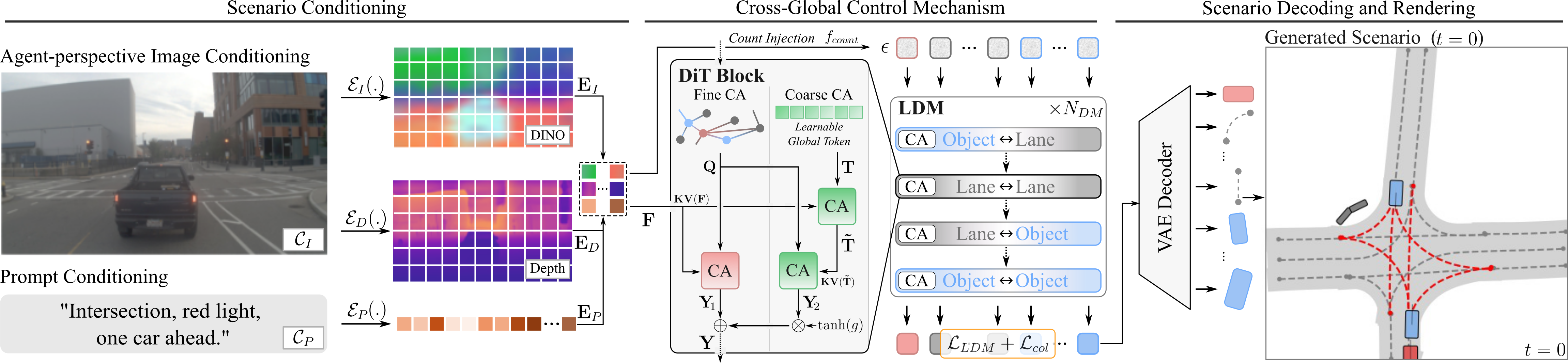}
\caption{\textbf{ScenarioControl Conditioning and Initial Scene Generation.} Our method generates controllable, vectorized driving scenarios from visual (option 1) or prompt (option 2) conditioning. Text prompt or image embeddings guide the latent diffusion model (LDM) via cross-global control mechanism to ensure structurally valid scene generation (right). Red, blue, and grey tinting in both the latent and decoded spaces indicate ego vehicle, other vehicles, and lanes, respectively.} 
\label{fig:method_overview}
\end{figure*}

The field of synthetic data generation for autonomous driving tasks can be clustered into \textit{traffic simulation}, \textit{scenario generation} methods and \textit{sensor data generation} methods. The former focuses on behavioral simulation of traffic participants, whereas the latter encompasses the generation of maps, agent start locations, and static road obstacles. The last one builds on established behavior rollouts and scenario layouts to generate corresponding multi-modal sensor\nolinebreak\ data. 

\PAR{Traffic Simulation.}
Traditional autonomous driving simulators such as CARLA\cite{dosovitskiy2017carlaopenurbandriving} and others~\cite{panpan2020summit, rong2020lgsvl} provide reproducible testbeds but rely on hand-crafted rules and scripted behaviors, limiting their ability to capture real-world driving diversity. Procedural generation approaches, including MetaDrive and SMARTS~\cite{li2022metadrivecomposingdiversedriving, zhou2020smartsscalablemultiagentreinforcement}, improve scalability but struggle with behavioral realism and rare event coverage.
Data-driven simulators address these limitations by learning directly from real-world logs. Systems such as Waymax and GPUDrive~\cite{gulino2023waymaxaccelerateddatadrivensimulator, kazemkhani2025gpudrivedatadrivenmultiagentdriving} enable hardware-accelerated training on logged scenarios, improving behavioral fidelity. However, replay-based approaches remain fundamentally constrained to observed patterns and cannot synthesize novel situations or explore counterfactual futures.

This limitation motivates generative simulators that synthesize new diverse and realistic scenarios, with several works addressing the behavior gap of log replay~\cite{rowe2024ctrlsim, feng2023trafficgen}. TrafficGen~\cite{feng2023trafficgen} generates both initial agent states and agent behavior, CtRL-Sim~\cite{rowe2024ctrlsim} specializes in the latter. They use a transformer-based driving policy to enable controllable and adversarial agent behaviors but assume pre-existing scene layouts. 

\PAR{Scenario Generation.}
Most traffic simulators require an initialized scene from which to roll out agent behaviors. A significant body of work focuses on producing suitable initial representations for agents~\cite{meng2025scenegen, scenecontrol2024}, lane graphs~\cite{hdmapgen2021} or both~\cite{chitta2024sledgesynthesizingdrivingenvironments, sun2024drivescenegen, rowe2025scenariodreamervectorizedlatent}, providing explicit structural control. When combined with traffic simulators, such initial scenes enable complete scenario generation with agent behaviors~\cite{chitta2024sledgesynthesizingdrivingenvironments, rowe2025scenariodreamervectorizedlatent}.
SLEDGE~\cite{chitta2024sledgesynthesizingdrivingenvironments} uses a raster-to-vector autoencoder with diffusion models to generate lane graphs and initial agent placements, but relies on rule-based traffic models for agent behaviors, limiting behavioral realism, while \cite{rowe2025scenariodreamervectorizedlatent} builds on a vectorized scene representation to both accelerate the generation and increase quality.
However, these methods do not allow for visual/textual grounding and focus on vectorized layouts for planners under flat-ground assumptions, rather than generating realistic sensor data.

\PAR{Sensor Data Generation.}
A line of work focuses on reconstructing photorealistic sensor observations. Neural rendering methods based on NeRF and 3D Gaussian Splatting, including UniSim, NeuRAD, and Street Gaussians~\cite{yang2023unisimneuralclosedloopsensor, tonderski2024neuradneuralrenderingautonomous, yan2024streetgaussiansmodelingdynamic}, achieve high visual fidelity but require explicit scene reconstruction and remain tethered to captured layouts, limiting their ability to generate counterfactual scenarios.
Diffusion-based world models offer a more flexible alternative by directly synthesizing sensor videos. GAIA-1~\cite{hu2023gaia1generativeworldmodel} demonstrates controllable multi-camera generation, while subsequent work scales to longer horizons (LongDWM) and improves multi-view consistency (GAIA-2)~\cite{wang2025longdwm, russell2025gaia2controllablemultiviewgenerative}. Cosmos-Drive-Dreams~\cite{nvidia2025cosmosdrivedreams} extends these capabilities by post-training the Cosmos world foundation model on large-scale driving data, enabling precise HDMap and 3D bounding box control, multi-view expansion from single views, and LiDAR point cloud generation alongside RGB synthesis. DriveArena~\cite{yang2024drivearenaclosedloopgenerativesimulation} explores closed-loop evaluation with diffusion models. Methods such as SimGen, GeoDrive and UMGen~\cite{zhou2024simgen, chen2025geodrive3dgeometryinformeddriving, umgen2025wu} improve controllability by conditioning image generation on structural layouts and 3D geometry though these methods typically focus on rasterized map or non-structured scene representations.

Methods that enable trajectory-level control, such as Epona~\cite{zhang2025epona} and ProphetDWM~\cite{wang2025prophetdwm}, or safety-critical scenario synthesis like AdvDiffuser~\cite{chen2023advdiffuser}, still operate either purely in abstract action control modes \cite{gao2024vistageneralizabledrivingworld} or purely in pixel space.

\vspace*{6pt}

\noindent
ScenarioControl bridges this gap and exposes explicit control handles that allow for a controllable scenario generation with descriptive, interpretable scene prompts, subsequent behavioral roll-outs, and final sensor data generation. Unlike existing vectorized methods that generate scenes unconditionally, we enable the conditional generation of initial scenes. This produces complete vectorized scene graphs with explicit topology and agent placement that reflect real-world observations, or can support the generation of large-scale new datasets for specific scenarios using prompting. These structured representations can then drive traffic simulators for behavioral rollouts and serve as conditioning for photorealistic video generation, enabling full multi-modal scenario generation that maintains both structural consistency and visual realism. Where diffusion video models provide visual diversity without structured guarantees and vectorized simulators provide structural control without visual grounding, we offer both: controllable structured synthesis realized as photorealistic generation.

\section{ScenarioControl}
\label{sec:method}
\begin{figure*}[t]
\centering
\includegraphics[width=1.0\linewidth]{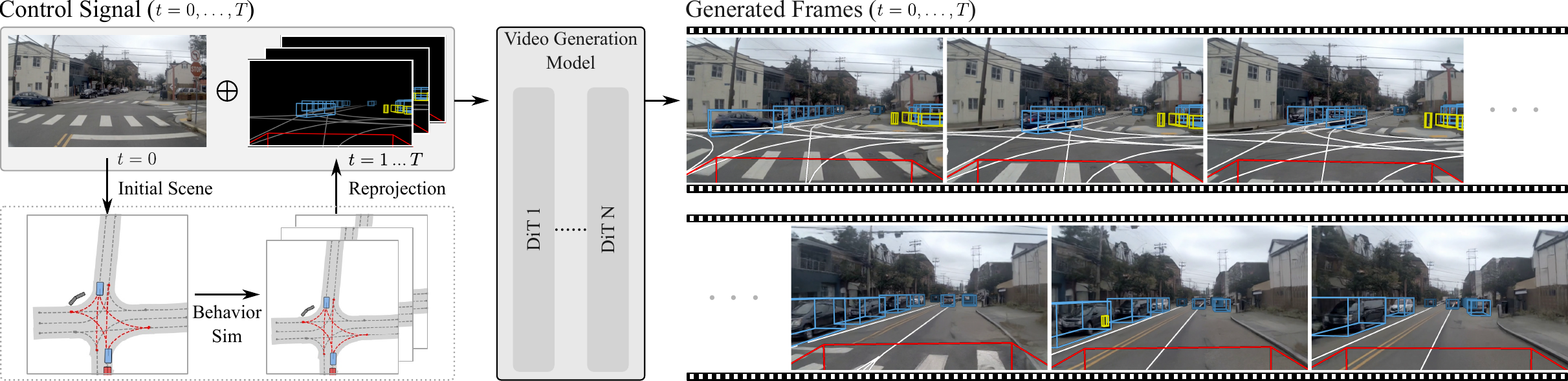}
\caption{\textbf{Scenario-Controlled Video Generation.} Our method produces driving scene representations ($t=0$) that we can simulate into scenario rollouts with a simulator, and use for downstream tasks such as video generation. The vectorized BEV scenario is projected into conditional camera-space layouts (BEV2Cam), and then fed into a LoRA-adapted Wan 2.2 (5B) model as a control signal for conditional video generation, together with the first frame.}

\label{fig:method_videogen}
\end{figure*}
In the following, we first describe our scene representation and vectorized scene generation method (\cref{ssec:rep}). Next, we describe the proposed conditioning mechanisms for camera observations or natural language text prompts (\cref{ssec:cldm}), also illustrated by Figure~\ref{fig:method_overview}. In~\cref{ssec:video-generation} and~\cref{fig:method_videogen}, we describe scenario-guided video generation through vision-language control. 

\subsection{Scene Representation and Generation}\label{ssec:rep}
We generate the initial scene by jointly modeling the underlying map structure and the actors’ initial states in a bird’s-eye-view (BEV) representation with elevation information. Specifically, we generate each scene within a $64\mathrm{m}\times 64\mathrm{m}$ field of view. A scene is then defined as a graph $\mathcal{I} = \{ \mathcal{O}, \mathcal{M} \}$, comprising a set of objects $\mathcal{O}$ and a map structure $\mathcal{M}=\{\mathcal{L}, \mathbf{A}\}$. The map structure consists lane centerlines $\mathcal{L}$ and their connectivity graph $\mathbf{A}$. Our objective is to sample a scene $\mathcal{I} \sim p(\,\cdot\,| \mathcal{C})$ from some distribution $p$, conditioned on inputs $\mathcal{C}$, which can be either an image captured from the perspective of an agent in the scene or a text prompt describing the scene.

In contrast to existing scenario diffusion approaches that abstract the environment in a purely two-dimensional BEV representation~\cite{rowe2025scenariodreamervectorizedlatent, chitta2024sledgesynthesizingdrivingenvironments}, we augment the object parametrization with vertical structure by incorporating elevation $z$ and object height $h$. This  provides essential elevation cues for downstream camera and sensor-data synthesis, facilitating faithful reprojection of generated scenarios into the image domain as shown in  Fig.~\ref{fig:video_generation_samples}. 

\subsection{ScenarioControl for Vectorized Latent Diffusion.}
\label{ssec:cldm}

We propose a controllable vectorized latent diffusion model, as illustrated in Fig.~\ref{fig:method_overview}, where we introduce the \emph{cross-global control mechanism} that fuses sparse vector tokens with dense attention-based conditioning from prompts and images.

First, we train a transformer-based vectorized autoencoder that encodes scene elements into a compact latent representation. 
We then train a diffusion model with $\epsilon$-predictor $\epsilon_\theta$ over the latents of the autoencoder $\mathbf{Z}=[\mathbf{Z}_{\mathcal{O}}, \mathbf{Z}_{\mathcal{L}}]$ with variable cardinalities $(N_o,N_l)$.
With condition $\mathcal{C}$,
we minimize the DDPM $\epsilon$-prediction objective~\cite{ho2020ddpm} with
\begin{equation}
\label{eq:loss_ldm}
\mathcal{L}_{\text{LDM}}
= \mathbb{E}_{\tau,\,\boldsymbol{\epsilon}\sim\mathcal{N}(0,\mathbf{I}),\,\mathcal{C}}
\big\|\boldsymbol{\epsilon} - \epsilon_\theta(\mathbf{Z}_\tau, \tau;\, \mathcal{C})\big\|_2^2,
\end{equation}
where $\mathbf{Z}_\tau$ are noisy latents at step $\tau$, composed of $N_o$ object latents and $N_l$ lane latents, and the noise vector decomposes as $\boldsymbol{\epsilon}=(\boldsymbol{\epsilon}_{\mathcal{O}},\boldsymbol{\epsilon}_{\mathcal{L}})$. 
The conditioning inputs $\mathcal{C}$ comprise dense control signals $\mathcal{C}_I$ (\emph{image}) and $\mathcal{C}_P$ (\emph{prompt}) used in cross-attention, and the default control tokens $\bar{c}$ that encode the number of agents and lanes $N=(N_o,N_l)$ and the scene/domain label $\mathcal{s}$ (e.g., Singapore, Las Vegas, Boston, Pittsburgh in nuPlan), injected via AdaLN-Zero conditioning~\cite{peebles2023scalablediffusionmodelstransformers}.

\vspace{\baselineskip}
\noindent
We next describe the proposed multi-modal control mechanism in more detail, which supports conditioning from both scene prompts and agent-perspective images. 

\vspace{5pt}
\noindent
\textit{Prompt Conditioning.}
\label{sec:text}
Given a prompt describing the present scene and actors, we employ a text encoder $\mathcal{E}_P$ that extracts token embeddings as 
\begin{equation}
    \mathbf{E}_P = \mathcal{E}_P(\mathcal{C}_P) \in \mathbb{R}^{M_P \times d_P},
\end{equation}
where $M_P$ is the number of prompt tokens and $d_P$ is the embedding dimension of the text encoder.  
The embeddings $\mathbf{E}_P$ are then projected through an MLP layer with the \emph{linear projection weights}  $\mathbf{W}_P$ to obtain control tokens $\mathbf{F}_P \in \mathbb{R}^{M_P \times d}$, where $d$ denotes the latent dimension used for agent or lane representations
\begin{equation}
    \mathbf{F}_P = \mathbf{E}_P \mathbf{W}_P.
\end{equation}
These prompt control tokens enter the control mechanism/attention by providing keys/values and are combined with the self-attention outputs. 

\vspace{5pt}
\noindent
\textit{Agent-Perspective Image Conditioning.}
\label{sec:image}
Given a forward-facing agent-perspective image $\mathcal{C}_I$, e.g., a dashcam style image, we extract dense features with a vision backbone $\mathcal{E}_{I}$ and depth estimator $\mathcal{E}_{D}$ as
\begin{equation}
\label{eq:image_features_extraction}
\begin{aligned}
\mathbf{E}_{\text{feat}} &= \mathcal{E}_{I}(\mathcal{C}_I)\in\mathbb{R}^{M_I\times d_{\text{feat}}},\\
\mathbf{E}_{\text{depth}} &= \mathcal{E}_{D}(\mathcal{C}_I)\in\mathbb{R}^{M_I\times d_{\text{depth}}},
\end{aligned}
\end{equation}
where $M_I$ denotes the number of image tokens, $d_{\text{feat}}$ the feature dimension of the vision backbone, and $d_{\text{depth}}$ the dimension of the estimated depth map. We use pretrained models for both the vision backbone and the depth estimator, and both are frozen during training.
Both image features and depth maps are projected to the model’s hidden dimension $d$ via learned linear mappings. Further, we add a non-trainable sine-cosine positional embedding $\mathbf{P}$ to the image features~\cite{peebles2023scalablediffusionmodelstransformers} and depth maps, enforcing a shared spatial encoding across modalities

\begin{equation}
\label{eq:modality_projection_posemb}
\mathbf{F}_{m} = \mathbf{E}_{m}\mathbf{W}_{m} + \mathbf{P},
\qquad m \in \{\text{feat}, \text{depth}\}.
\end{equation}
where $\mathbf{W}_{\text{feat}}$ and $\mathbf{W}_{\text{depth}}$ are learned \emph{linear projection weights} mapping modality-specific features to the shared hidden dimension $d$, and $\mathbf{P}$ ensures alignment within the same image coordinate frame across both image features and depth maps. 
Finally, we concatenate these position-aware tokens to form the control feature representation $\mathbf{F}_I = [\mathbf{F}_{\text{feat}}, \mathbf{F}_{\text{depth}}]$.\\

\vspace{5pt}
\noindent
\textit{Cross-Global Control Mechanism.}
Given the conditioning features, $\mathbf{F}_I$ and $\mathbf{F}_P$, we introduce a control mechanism to steer scenario generation. We employ $N_{\text{DM}}$ factorized attention blocks, each applying (in order) object-to-lane, lane-to-lane, lane-to-object, and object-to-object self-attention (SA) over the stacked object and lane tokens. The conditioning inputs are injected into each (CA) component, which we detail in the following section. An overview of the full mechanism is illustrated in Fig.~\ref{fig:method_overview}.

Conditioning vectorized scene tokens (lanes and agents) on agent-perspective images or scene prompt embeddings is inherently \emph{unaligned}: a single scene token may depend on evidence from arbitrary subsets of conditioning tokens (e.g., occluded actors, distant lane cues, or globally specified textual constraints). Although cross-attention can, in principle, model such dependencies by allowing each query to attend to all keys, it provides little inductive structure and can be sample-inefficient when learning global context.

We therefore compute cross-attention through two parallel branches that share the key/value projections of the conditioning stream. Given scene queries $\mathbf{Q}\in\mathbb{R}^{N_q\times d}$ and conditioning tokens $\mathbf{F}\in\mathbb{R}^{N_k\times d}$, we first compute cross-attention
\begin{equation}
\mathbf{Y}_{1} \;=\; \mathrm{Attn}\!\left(\mathbf{Q},\, \mathbf{K}(\mathbf{F}),\, \mathbf{V}(\mathbf{F})\right),
\end{equation}
implemented efficiently with FlashAttention. In parallel, we introduce a small set of learned latent tokens $\mathbf{T}\in\mathbb{R}^{L\times d}$ that aggregate global context from $\mathbf{F}$, and expose this compact summary back to the scene queries 
\begin{equation}
\mathbf{Y}_{2} \;=\; \mathrm{Attn}\!\left(\mathbf{Q},\, \mathbf{K}(\tilde{\mathbf{T}}),\, \mathbf{V}(\tilde{\mathbf{T}})\right), \;
\end{equation}
with
\begin{equation}
\quad \tilde{\mathbf{T}} \;=\; \mathrm{Attn}\!\left(\mathbf{T},\, \mathbf{K}(\mathbf{F}),\, \mathbf{V}(\mathbf{F})\right),
\end{equation}
with cost $\mathcal{O}(L N_k + N_q L)$. We combine both branches 
$\mathbf{Y} = \mathbf{Y}_{1} + \text{tanh}(g)\,\mathbf{Y}_{2}$ with a learned gate $g$. The gate is initialized such that $\tanh(g) \approx 0$ (i.e., $g = 0$), ensuring that the module initially reduces exactly to standard cross-attention. During training, the model can then progressively improve by selectively incorporating the additional global-context pathway. Since both pathways share the $\mathbf{K}(\cdot)$ and $\mathbf{V}(\cdot)$ projections, the parameter overhead remains minimal. Finally, the output is fused with multi-head self-attention via AdaLN-Zero modulation.

\vspace{5pt}
\noindent
\textit{Count Injection $f_{count}$}. Graph-based representations offer a natural handle for controlling scene complexity: the number of lanes and agents can be set directly by initializing the corresponding numbers of lane and object nodes. For instance, generating a two-lane highway can be guided by instantiating the matching number of lane nodes, while in the image/prompt-conditioned setting, these counts can also be inferred from the conditioning signal. We therefore train a lightweight attention-based regressor $f_{\text{count}}$ that predicts the number of agents and lanes $(N_o, N_l)$ from the conditioning tokens $\mathbf{F}$.

We note that compared to raster encodings or post-hoc control modules (e.g., ControlNet/T2I-Adapter~\cite{zhang2023controlnet,mou2023t2iadapterlearningadaptersdig}), our method directly operates on variable-length vector tokens, thereby preserving topology (lane connectivity) and scene structure.

\begin{figure*}[t]
    \centering
    \includegraphics[width=\linewidth]{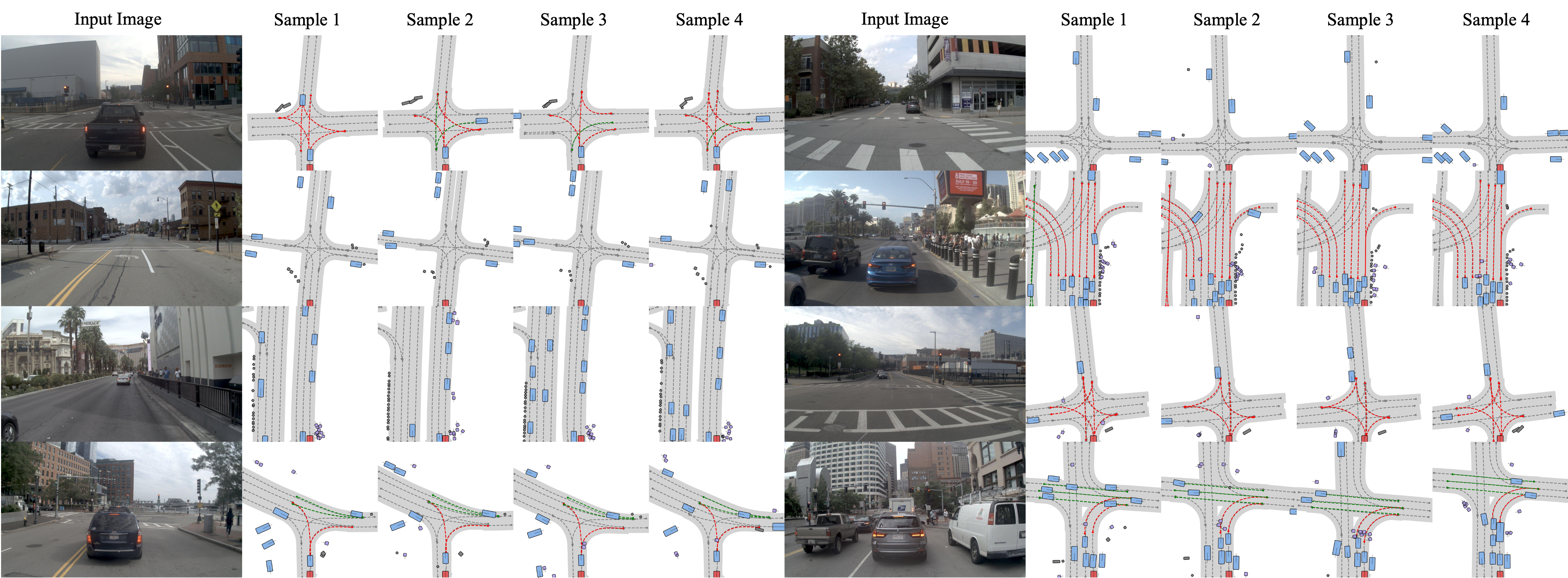}
    \caption{\textbf{Dashcam-Style Image-Conditioned Scene Generation}. Our model generates diverse initial scenes across different diffusion samples given a single input image. Elements with high certainty, such as vehicles in visible range, are found consistently across most samples. Elements with lower certainty, e.g., far away or outside the FOV, are sampled plausibly by our model.}
    \label{fig:exp img}
    \vspace{-4mm}
\end{figure*}

\subsection{Video Generation and Continuation}
\label{ssec:video-generation}

The use of vectorized scene graph representation $\mathcal{I}$ allows for direct BEV behavior simulation (with elevation) and camera sensor video synthesis without requiring an additional learned lifting step, as is the case for rasterized representations. For behavior simulation, the vectorized representation is used directly, while for video generation it is projected into the camera coordinate frame to create control inputs for a video diffusion model, both described in detail in the following.

\PAR{Behavior Simulation.} Real-world cameras capture far beyond the 64m of an initial scenario layout. We use diffusion outpainting to construct large-scale scenes, which are then temporally rolled out using a behavior model $\mathcal{B}_T$ to produce diverse and behaviorally consistent agent trajectories. From this, we obtain scenario rollout representations $\{\mathcal{I}_{t}\}_{t=1}^{T}$, which we reproject from BEV to camera space as wireframe sequences $\{\hat{\mathcal{C}}_{\text{wire},t}\}_{t=1}^T$. 

\PAR{Sensor Video Generation.} 
We adapt a video generation model $\mathcal{V}_i$ to generate photorealistic clips from the behavior simulation. We train two variants: an image-conditioned model that uses the first frame $\mathcal{C}_I$ for appearance, and a prompt-conditioned model where the scene appearance is controlled by prompt description. Both are also trained to adhere to control sequences (wireframe renders of the behavior simulation) $\{\hat{\mathcal{C}}_{\text{wire},t}\}_{t=1}^T$ to obtain photorealistic multi-frame renders $\{\hat{I}_{t}\}_{t=0}^T$ that follow the simulated agent behavior (see Figure \ref{fig:method_videogen}). Since we ground the video generation with the vectorized scene representation, we can transform the camera pose and generate videos from the perspectives of agents other than the ego agent whose camera capture was used to initialize the scene. We do so by re-rendering the wireframe representation with different camera extrinsics while maintaining a consistent text prompt.

\begin{figure*}[t]
    \centering
    \includegraphics[width=0.99\linewidth]{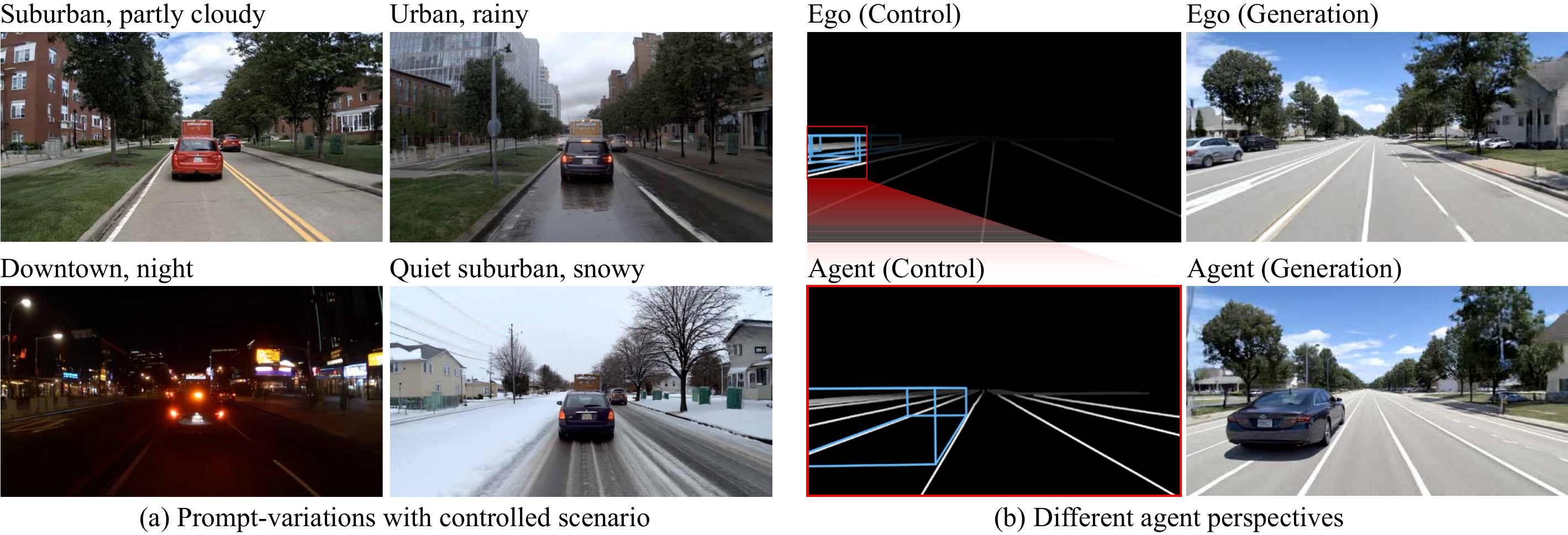}
    \caption{\textbf{Prompt-conditioned Scenario Adaptations.} 
    The prompt-conditioned video generation variant allows (a) different scene variants given the same generated traffic rollout, and (b) generating the same traffic situation from different perspectives by transforming the camera pose while keeping the text prompt constant.}
    \label{fig:video_generation_samples_text}
\end{figure*}

After adaptation, the model achieves fine-grained control of traffic behavior -- either as part of a video continuation task with the first-frame conditioning or in completely novel traffic situations from a prompt. Examples of controlled rollouts are reported in Figures \ref{fig:video_generation_samples} and \ref{fig:video_generation_samples_text}, respectively.

\begin{figure*}[t]
    \centering
    \includegraphics[width=\linewidth]{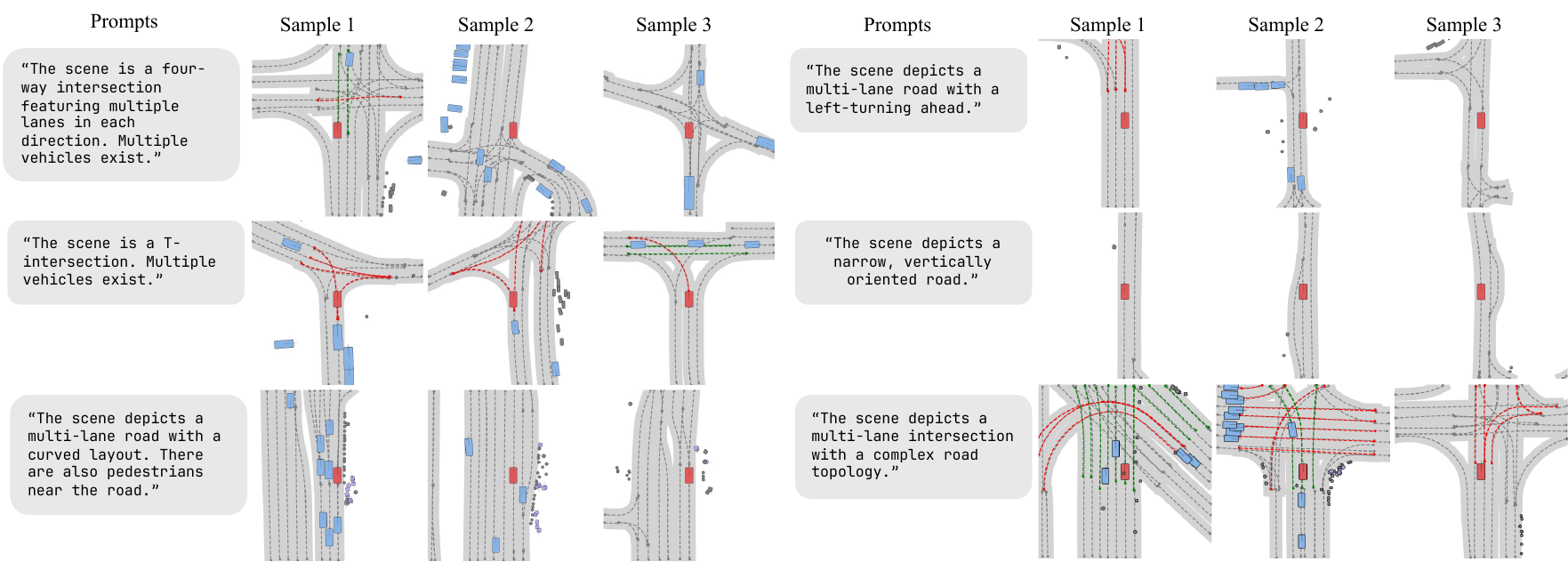}
    \caption{\textbf{Prompt-Controlled Scenario Generation}. Our model generates diverse initial scenes across different diffusion samples, with lane and object placement in each adhering to the text prompt. Text prompts are more open-ended than images, resulting in a larger variety across samples.
    }
    \label{fig:exp text}
    \vspace{-1eM}
\end{figure*}

\subsection{Training}
\label{ssec: training and sampling}

We adopt a two-stage training strategy. We first train the encoder $f_e$ and decoder $f_d$ to learn a reliable projection into the latent space. Next, we train the generative control mechanism and freeze the $f_e,f_d$ weights and only train the dense condition blocks as described in \cref{ssec:cldm}, leveraging either the condition on Prompt or Image conditions. 

Additionally, we also employ \emph{classifier-free guidance} (CFG)~\cite{ho2022classifierfreediffusionguidance}. Specifically, the conditioning inputs $\mathbf{F}_I$ and $\mathbf{F}_P$  are randomly dropped with probability $p_{\text{CFG}}$ during training,  encouraging the model to learn both conditional and unconditional behaviors.
At inference, the guided prediction is computed as
\begin{equation}
\epsilon_{\mathrm{CFG}} = 
\epsilon_\theta(\mathbf{Z}_\tau, \tau; \varnothing)
+ w \big[
\epsilon_\theta(\mathbf{Z}_\tau, \tau; \mathcal{C})
- \epsilon_\theta(\mathbf{Z}_\tau, \tau; \varnothing)
\big],
\end{equation}
where $w$ is the guidance weight controlling the strength of conditioning.

\PAR{Scene Layouts.}
At test time, we support three generation modes that reflect the asymmetry between prompt and image conditioning. Text prompts can describe the full surrounding around the ego agent, whereas a single camera frame only constrains the visible region in front of the ego. We therefore define two canonical scene layouts with dimension of 64x64m: 1. an ego-centered crop $\mathcal{F}_P$ and 2. a forward-only crop $\mathcal{F}_I$, obtained by shifting the crop such that the ego lies near the edge, to maximize coverage ahead.

Crucially, we train the model on the same layout types used at inference, enabling consistent long-horizon rollouts. In practice, we use prompt-controlled synthesis for $\mathcal{F}_P$, image-conditioned completion for $\mathcal{F}_I$, and forward outpainting to extend either crop beyond the current field-of-view by sampling additional scene-graph nodes. Completion and outpainting are realized with masked denoising: latents corresponding to observed nodes are clamped, while the remaining tokens are sampled conditioned on $\mathcal{C}$.

\PAR{Collision Penalty.}
In practice, conditioning on prompt and images can induce \emph{cluttered} scene hypotheses: images contain occlusions and missing context, while prompt is often underspecified. Both effects can place multiple agents into the same plausible region, resulting in overlapping boxes in the decoded scene graph. To encourage physically consistent layouts, we add a collision penalty $\mathcal{L}_{\text{col}}$ during training. Concretely, we decode intermediate latents at selected timesteps and compute pairwise $(i,j)$ overlaps ($\text{overlap}(i,j)$) between predicted agents $i,j$ with an intersection-over-union of their corresponding bounding boxes. We define the collision regularization loss as
\begin{equation}
\mathcal{L}_{\text{col}} \;=\; \frac{1}{N}\sum_{i\neq j} \tanh\!\left(\frac{\text{overlap}(i,j)}{\zeta}\right),
\end{equation}
where $\zeta$ controls smoothness. Since decoding is unreliable at low signal-to-noise ratios, we weight the penalty by
$w_\tau = 1 - \sqrt{1 - \bar{\alpha}_\tau}$, emphasizing later diffusion steps (smaller $\tau$) where the predicted geometry is more meaningful.
This regularizer reduces agent overlap in the initial scene and improves global scene consistency under ambiguous conditioning. 

\PAR{Implementation Details.} Further details on scene definitions, model architecture, and training and inference procedures are provided in the appendix.
\section{Vision-language Scenario Dataset}
\label{ssec: dataset}
\begin{figure*}[t]
    \centering
    \includegraphics[width=0.99\linewidth]{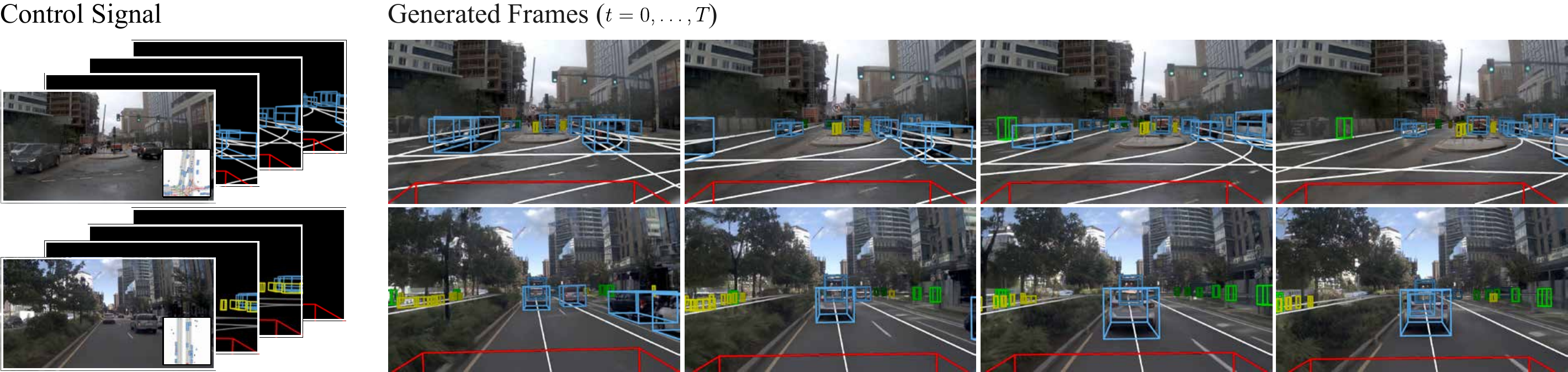}
    \caption{\textbf{Controllable Video Continuations.} 
    Our video generation model respects control signals from the reprojected wireframe image sequences, and remains visually consistent with the initial frame. To show adherence with the control signal, we overlay the wireframe control signals over the generated sequences for qualitative evaluation. The samples shown are drawn from the NuPlan test split.}
    \label{fig:video_generation_samples}
    \vspace{-3mm}
\end{figure*}

To facilitate the training of our vision and language conditioned model, we curate a dataset consisting of images, natural language descriptions, and BEV maps.
We select driving scenarios with corresponding camera captures from the nuPlan dataset~\cite{Karnchanachari2024TowardsLPnuplan}. To generate scene-level captions (e.g., “An intersection with a red light and multiple vehicles on the road. Pedestrians are standing on the sidewalk.”), we first render BEV visualization images for each scene and use a VLM (GPT-4.1-mini) to produce descriptive captions. 

This process results in a large-scale dataset containing approximately 500K  high-quality captions, providing a rich foundation for training and evaluating our multimodal model. We provide additional details and dataset samples in the Appendix.

\section{Experiments}
\label{sec:experiements}
Next, we first introduce the relevant evaluation metrics in Sec.~\ref{sec:metrics}. We then compare our proposed Cross-Global Control conditioning mechanism with other popular conditioning mechanisms in Sec.~\ref{sec:ControlAblation}. Subsequently, we demonstrate superior adherence to control input with significant overlap in scene content in Sec.~\ref{sec:ControlAdherence}. Next, we analyze the inclusion to predict the actor and lane counts, $f_{count}$, and to suppress collisions through the loss $\mathcal{L}_{\text{col}}$.
Lastly, we provide a generalization experiment on the Waymo motion dataset \cite{Sun_2020_CVPR_waymo_perception} in Sec.~\ref{sec:Generalization}. As our method focuses on \emph{controllable} scenario generation, unconditioned comparison with prior methods \cite{chitta2024sledgesynthesizingdrivingenvironments, rowe2025scenariodreamervectorizedlatent,feng2023trafficgen} are orthogonal but can still be found in the Appendix. 

\setlength{\tabcolsep}{12pt}   
\begin{table*}[t!]
\centering
\caption{\textbf{Controllability Evaluation on Nuplan.}
We evaluate controllability with three complementary metrics: cosine control similarity (CCS), shuffled perturbation gap (SPG), and control sensitivity correlation (CSC). Across both image- and text-conditioned settings, our method consistently improves all three scores, indicating stronger adherence to the specified controls $\mathcal{F}_i$.}
\label{tab:control_metrics_methods}
\resizebox{0.85\linewidth}{!}{
\setlength{\tabcolsep}{4pt}
\renewcommand{\arraystretch}{1}
\begin{tabular}{ll ccc ccc ccc}
    \toprule
    \multirow{2}{*}{\textbf{$\mathcal{F}_i$}} & \multirow{2}{*}{\textbf{Method}} 
    & \multicolumn{3}{c}{\textbf{Global Control}} 
    & \multicolumn{3}{c}{\textbf{Lane Control}} 
    & \multicolumn{3}{c}{\textbf{Agent Control}} \\
    \cmidrule(lr){3-5} \cmidrule(lr){6-8} \cmidrule(lr){9-11}
    & & CCS $\uparrow$ & SPG $\uparrow$ & CSC $\uparrow$
     & CCS $\uparrow$ & SPG $\uparrow$ & CSC $\uparrow$
     & CCS $\uparrow$ & SPG $\uparrow$ & CSC $\uparrow$ \\
    \midrule
    \multirow{2}{*}{\rot{$i=P$}}&  Scenario Dreamer~\cite{rowe2025scenariodreamervectorizedlatent} & 0.9896 & 0.0082 & 0.3510 
                  & 0.4074 & 0.3568 & 0.2885
                  & 0.1909 & 0.1732 & 0.0625 \\
     & ScenarioControl & \textbf{0.9910} & \textbf{0.0097} & \textbf{0.3764} 
                  & \textbf{0.4752} & \textbf{0.4335} & \textbf{0.3855} 
                  & \textbf{0.2992} & \textbf{0.3024} & \textbf{0.1499} \\
    \cmidrule(lr){1-11}
    \multirow{2}{*}{\rot{$i=I$}}& Scenario Dreamer~\cite{rowe2025scenariodreamervectorizedlatent}
      & 0.9887 & 0.0141 & 0.4320 & 0.4400 & 0.3965 & 0.3166 & 0.1868 & 0.1800 & 0.0623 \\
     & ScenarioControl & \textbf{0.9926} & \textbf{0.0194} & \textbf{0.6274}
        & \textbf{0.6606} & \textbf{0.6108} & \textbf{0.6100}
        & \textbf{0.3853} & \textbf{0.3865} & \textbf{0.2255}\\
    \bottomrule
\end{tabular}}
\vspace{0.8eM}
\caption{\textbf{Controllability Evaluation on the Waymo Motion dataset.} We evaluate controllability on prompt-conditioned settings on Waymo, indicating strong cross-dataset generalization.}\label{tab:WaymoGeneralization}
\resizebox{0.85\linewidth}{!}{
\centering
\setlength{\tabcolsep}{4pt}
\renewcommand{\arraystretch}{1}
\begin{tabular}{ll ccc ccc ccc}
    \toprule
    \multirow{2}{*}{\textbf{$\mathcal{F}_i$}} & \multirow{2}{*}{\textbf{Method}} 
    & \multicolumn{3}{c}{\textbf{Global Control}} 
    & \multicolumn{3}{c}{\textbf{Lane Control}} 
    & \multicolumn{3}{c}{\textbf{Agent Control}} \\
    \cmidrule(lr){3-5} \cmidrule(lr){6-8} \cmidrule(lr){9-11}
     & & CCS $\uparrow$ & SPG $\uparrow$ & CSC $\uparrow$
     & CCS $\uparrow$ & SPG $\uparrow$ & CSC $\uparrow$
     & CCS $\uparrow$ & SPG $\uparrow$ & CSC $\uparrow$ \\

    \midrule
    \multirow{2}{*}{\rot{$i=P$}}& Scenario Dreamer~\cite{rowe2025scenariodreamervectorizedlatent} 
        & 0.9922 & 0.0123 & 0.6701 & 0.3278 & 0.2775 & 0.2128 & 0.2264 & 0.2084 & 0.1220 \\
    &ScenarioControl
        & \textbf{0.9950} & \textbf{0.0171} & \textbf{0.7587} & \textbf{0.6739} & \textbf{0.6204} & \textbf{0.5179} & \textbf{0.3586} & \textbf{0.3529} & \textbf{0.2112} \\
    \bottomrule    
\end{tabular}}
\end{table*}

\setlength{\tabcolsep}{12pt}   
\begin{table*}[t!]
\centering
\caption{\textbf{Analysis of Cross-Global Control Mechanism.}
We analyze the effect of our cross-global control mechanism with three complementary metrics: cosine control similarity (CCS), shuffled perturbation gap (SPG), and control sensitivity correlation (CSC). Across both image- and prompt-conditioned settings, our method consistently improves all three scores, indicating stronger adherence to the specified controls $\mathcal{F}_i$.}
\label{tab:image_conditioning_controllability}
\vspace{-1eM}

\resizebox{0.99\linewidth}{!}{
\setlength{\tabcolsep}{4pt}
\renewcommand{\arraystretch}{1}
\begin{tabular}{ll ccc ccc ccc cc}
    \toprule
    \multirow{2}{*}{\textbf{$\mathcal{F}_i$}} & \multirow{2}{*}{\textbf{Method}}
    & \multicolumn{3}{c}{\textbf{Global Control}}
    & \multicolumn{3}{c}{\textbf{Lane Control}}
    & \multicolumn{3}{c}{\textbf{Agent Control}}
    & \multirow{2}{*}{\textbf{Agent}}
    & \multirow{2}{*}{\textbf{Collision}} \\
    \cmidrule(lr){3-5} \cmidrule(lr){6-8} \cmidrule(lr){9-11}
    & & CCS $\uparrow$ & SPG $\uparrow$ & CSC $\uparrow$
      & CCS $\uparrow$ & SPG $\uparrow$ & CSC $\uparrow$
      & CCS $\uparrow$ & SPG $\uparrow$ & CSC $\uparrow$
      & AP $\uparrow$ & RATE $\downarrow$\\
    \midrule

    \multirow{7}{*}{\rot{ $i=P$}}

    & Concatenation
    & 0.9898 & 0.0089 & 0.3606 & 0.4768 & 0.4286 & 0.3845 & 0.2786 & 0.2693 & 0.1340 & 23.03 & 24.43 \\
    
      & Full Cross-Attention~\cite{AttentionISAllYouNeed}
      & 0.9903 & 0.0093 & 0.3638 & 0.4803 & 0.4274 & 0.3830 & 0.2762 & 0.2722 & 0.1347 
      & 22.54 & 21.56 \\
      
      & Gated Attention~\cite{Flamingo}
      & 0.9894 & 0.0082 & 0.3309 & 0.4013 & 0.3505 & 0.2695 & 0.1668 & 0.1624 & 0.0581 
      & 20.66 & 21.15 \\
      
      & Linear Attention~\cite{FlattenTransformer}
      & 0.9905 & 0.0086 & 0.3705 & 0.4416 & 0.3942 & 0.3505 & 0.2296 & 0.2303 & 0.0934 
      & 19.85 & \textbf{19.18} \\

      & AgentAttention~\cite{agentattention}
    & 0.9902 & 0.0091 & 0.3639 & 0.4827 & 0.4312 & 0.3893 & 0.2716 & 0.2667 & 0.1334 & 22.78  & 21.11 \\
    
    & SAAP Cross-Attention~\cite{mazare2025inference}
    & 0.9904 & 0.0096 & 0.3628 & 0.4821 & 0.4278 & 0.3722 & 0.2556 & 0.2562 & 0.1264 & 22.03 & 27.51 \\
      
      & Windowed Attention~\cite{FlashAttention2}
      & 0.9721 & 0.0078 & 0.1981 & 0.0899 & 0.1387 & 0.0893 & 0.0607 & 0.0655 & 0.0414 
      & 5.55 & 45.53 \\
      
      & Deformable Attention~\cite{DeformableDETR}
      & 0.9802 & 0.0081 & 0.2503 & 0.3678 & 0.3301 & 0.2385 & 0.1439 & 0.1463 & 0.0525 
      & 21.65 & 23.62 \\
      
      & Squeezed Attention~\cite{squeezedAttention}
      & 0.9882 & 0.0084 & 0.3404 & 0.4707 & 0.4202 & 0.3702 & 0.2021 & 0.2048 & 0.0869 
      & \textbf{23.97} & 20.00 \\
      
      & Cross-Global Control (Ours)
      & \textbf{0.9908} & \textbf{0.0098} & \textbf{0.3809} 
      & \textbf{0.4900} & \textbf{0.4367} & \textbf{0.4006} 
      & \textbf{0.3163} & \textbf{0.3096} & \textbf{0.1626} 
      & 23.36 & 20.07 \\
    
    \cmidrule(lr){1-13}

    \multirow{10}{*}{\rot{i = I}} 
      
      & Concatenation
        & 0.9881 & 0.0153 & 0.4105
        & 0.5564 & 0.5116 & 0.4094
        & 0.2501 & 0.2537 & 0.1314
        & 23.37 & 30.18 \\
        
        & Full Cross-Attention~\cite{AttentionISAllYouNeed}
        & 0.9924 & 0.0191 & 0.6137
        & 0.6596 & 0.6098 & 0.6075
        & 0.3766 & 0.3788 & 0.2195
        & 32.00 & 16.07 \\
        
        & Gated Attention~\cite{Flamingo}
        & 0.9896 & 0.0156 & 0.4637
        & 0.5795 & 0.5315 & 0.4953
        & 0.2958 & 0.2945 & 0.1514
        & 25.13 & 14.92 \\
        
        & Linear Attention~\cite{FlattenTransformer}
        & 0.9901 & 0.0177 & 0.5299
        & 0.6390 & 0.5899 & 0.5861
        & 0.3527 & 0.3500 & 0.2012
        & 27.20 & 16.92 \\
        
        & AgentAttention~\cite{agentattention}
        & 0.9895 & 0.0159 & 0.4729
        & 0.5963 & 0.5478 & 0.5110
        & 0.3032 & 0.3051 & 0.1632
        & 24.54 & 17.18 \\
        
        & SAAP Cross-Attention~\cite{mazare2025inference}
        & 0.9907 & 0.0180 & 0.5414
        & 0.5849 & 0.5336 & 0.4670
        & 0.2531 & 0.2700 & 0.1318
        & 26.05 & 18.17 \\
        
        & Windowed Attention~\cite{FlashAttention2}
        & 0.9902 & 0.0180 & 0.5467
        & 0.6356 & 0.5848 & 0.5644
        & 0.2880 & 0.2905 & 0.1450
        & 23.60 & 16.22 \\
        
        & Deformable Attention~\cite{DeformableDETR}
        & 0.9883 & 0.0151 & 0.4546 & 0.4799 & 0.4415 & 0.3194 & 0.2058 & 0.2155 & 0.1016 & 21.60 & \textbf{13.82} \\
        
        & Squeezed Attention~\cite{squeezedAttention}
        & 0.9901 & 0.0169 & 0.4927
        & 0.6183 & 0.5675 & 0.5423
        & 0.3167 & 0.3186 & 0.1689
        & 28.25 & 19.94 \\
        
        & Cross-Global Control (Ours)
        & \textbf{0.9926} & \textbf{0.0194} & \textbf{0.6274}
        & \textbf{0.6606} & \textbf{0.6108} & \textbf{0.6100}
        & \textbf{0.3853} & \textbf{0.3865} & \textbf{0.2255}
        & \textbf{33.46} & 17.07 \\
    \bottomrule
    
\end{tabular}
}
\vspace{-4mm}
\end{table*}

\subsection{Evaluation Metrics}\label{sec:metrics}
Our evaluation focuses on two key aspects: generation quality and controllability. For generation realism, we adopt the same lane- and agent-level metrics as~\cite{rowe2025scenariodreamervectorizedlatent} and are presented in the appendix. 
For controllability we introduce a set of control metrics to assess adherence to the provided conditioning signals that build on top of Agent Accuracy, Collision Rate and Control Adherence.

\noindent\textbf{Agent Accuracy.} As a first step, we match generated agents to ground-truth agents and evaluate placement accuracy using average precision (AP). We compute $\mathrm{AP}_{\delta}$ under center-point distance thresholds $\delta \in \{1.0,\,2.0,\,3.0\}\,\mathrm{m}$ and report their mean, yielding an equally weighted measure of how well the control signal localizes agents. For the image-conditioned setting, AP is evaluated only for agents within the camera field of view (FOV), whereas for the prompt-conditioned setting, AP is computed over all generated agents.

\noindent\textbf{Collision Rate.} We calculate the intersection over union over all predicted agents and report the fraction of scenarios with collisions in which actors and/or scene object bounding boxes intersect.

\noindent\textbf{Control Adherence.} To quantify how well the predicted global, lane, and agent attributes follow the intended controls, we report three complementary metrics: Cosine Control Similarity (CCS), Shuffled Perturbation Gap (SPG), and Control Sensitivity Correlation (CSC), which are detailed in the appendix.
\textbf{CCS} measures the alignment between the conditioning signal and the corresponding change in the generated scene, capturing direct controllability.
\textbf{SPG} evaluates causal dependence by comparing the model’s response to correct versus randomly shuffled conditions; a larger gap indicates stronger conditional consistency.
\textbf{CSC} measures the correlation between variations in the conditioning input and variations in the generated output, reflecting the sensitivity and smoothness of control.
We evaluate across three levels (global, lane, and agent) to separately assess control over large-scale layout, map structure, and agents.

\subsection{Analysis of Control Mechanism}\label{sec:ControlAblation}
\begin{figure*}[t!]
\includegraphics[width=\linewidth]{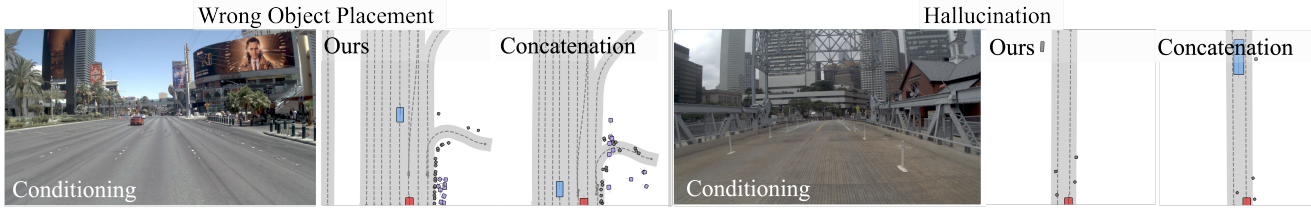}
\caption{\textbf{Qualitative comparison with Concatenation. }Concatenation performs worse across all metrics and frequently violates the control input, e.g., generating off-FOV content (left) or hallucinated agents (right).}\label{fig:qual_comparison_w_cocnat}
\vspace{-1eM}
\end{figure*}

We validate our cross-global control module by comparing it against a broad set of attention designs commonly used for fusing dense conditioning signals with token sequences. To our knowledge, this is the first study that systematically evaluates such mechanisms for conditioning a \emph{vectorized 3D scene graph} on dense prompt or image features. Specifically, we compare against simple concatenation, full cross-attention~\cite{AttentionISAllYouNeed}, gated attention~\cite{Flamingo}, linear attention~\cite{FlattenTransformer}, AgentAttention~\cite{agentattention}, SAAP cross-attention~\cite{mazare2025inference}, windowed attention~\cite{FlashAttention2}, deformable attention~\cite{DeformableDETR}, and squeezed attention~\cite{squeezedAttention}. 
Quantitative results on lane- and agent-control metrics are reported in Tab. \ref{tab:image_conditioning_controllability} with a qualitative result shown in Figure~\ref{fig:qual_comparison_w_cocnat}. 

Overall, attention-based conditioning mechanisms consistently outperform simple concatenation, and this trend holds for both prompt- and image-conditioned control. In the image-conditioned setting, our method reduces collision rate by $43.4\%$ and improves global CSC by $52.8\%$ over concatenation; moreover, it also improves over full cross-attention with a $2.7\%$ gain in agent CSC and a $4.6\%$ gain in AP, while also achieving stronger global and lane control. In the prompt-conditioned setting, our method also surpasses all other baselines on most controllability metrics. This confirms that our cross-global attention better captures the dense-to-sparse correspondences needed to align scene tokens with the specified controls $\mathcal{F}_i$.

\subsection{Analysis of Control Adherence}\label{sec:ControlAdherence}
Figures~\ref{fig:exp img} and~\ref{fig:exp text} illustrate controlled initial scenes generated with image and prompt conditioning, respectively: image inputs preserve visible structure and yield plausible completions of unobserved regions, while text prompts focus on key differentiators like intersection types allowing for more diverse sampling. 

Figure~\ref{fig:video_generation_samples} shows controllable video continuations that stay visually consistent with the initial view and follow the projected wireframe control signals over time. Finally, Table~\ref{tab:control_metrics_methods} reports higher CCS, SPG, and CSC across global, lane, and agent levels for the conditioned model, confirming improved control adherence and semantic alignment without compromising generation stability relative to the unconditioned model in \cite{rowe2025scenariodreamervectorizedlatent}.

\subsection{\texorpdfstring
  {Analysis of $f_{\text{count}}$ and $\mathcal{L}_{\mathrm{col}}$}
  {Analysis of f\_count and collision loss}}
\label{sec:AbalationStudy}
Table~\ref{tab:collision_loss} analyzes supervision for predicting the number of actors and lane elements from the conditioning signal via $f_{\text{count}}$, and for encouraging actor separation via the collision loss. The ablation study is conducted on a subset of the test set. Adding the collision loss consistently reduces collisions in the generated scenarios, regardless of whether $f_{\text{count}}$ is enabled. Using $f_{\text{count}}$ further lowers collision rate but also reduces AP, reflecting a trade-off: when the count is predicted rather than provided, the model tends to miss unseen or occluded actors outside the FOV. This reduces the number of placed agents, which decreases AP and, as a side effect, lowers the collision rate. 

\begin{table}[t!]  
\centering
\scriptsize
\setlength{\tabcolsep}{3pt} 
\caption{\textbf{Analysis of $f_{count}$ and $\mathcal{L}$}  
We report agent AP and simulated collision rates, while ablating the collision loss $\mathcal{L}_{col}$ and predicted agent counts $f_{count}$. Our final model significantly reduces collisions, with only a minor drop in AP.}
\label{tab:collision_loss}
\begin{tabular}{l cc cc}
    \toprule
        & \textbf{$\mathcal{L}_{col}$} 
        & \textbf{$f_{\text{count}}$}
        & \textbf{Collision}
        & \textbf{Agent} \\
        & 
        & 
        & RATE $\downarrow$ 
        & AP $\uparrow$  \\
    \midrule
        & \xmark & \xmark & 19.81 & 39.55  \\
        & \cmark & \xmark & 14.07 & 38.99 \\
        & \xmark & \cmark & 18.78 & 38.70 \\
        & \cmark & \cmark & 13.36 & 38.03 \\
    \bottomrule
\end{tabular}
\end{table}

\subsection{Generalization}\label{sec:Generalization}
We find that ScenarioControl generalizes across driving datasets to \cite{Sun_2020_CVPR_waymo_perception}. We evaluate transfer to the Waymo Open Motion dataset and report results in Tab.~\ref{tab:WaymoGeneralization}. By injecting explicit text prompts as control signals during generation, our model produces scenarios that better match the target distribution, outperforming Scenario Dreamer~\cite{rowe2025scenariodreamervectorizedlatent} across all reported metrics by up to \textit{143}\%.

\subsection{Comparisons to Agent Placement Techniques}
ScenarioControl generates both lane topology and
agent placements, producing truly novel scene realizations with a given conditioning signal.
It can also be applied in map-conditioned settings: by encoding a pre-existing lane topology from an existing map and denoising only the agent latents the
model places vehicles on a provided road topology without altering the
underlying geometry. This allows a direct comparison with methods such as
TrafficGen~\cite{feng2023trafficgen} that focus solely on agent placement given a fixed
map context. 

TrafficGen~\cite{feng2023trafficgen} is an autoregressive 
method that generates vehicle 
initial states conditioned on the road context without any vision-language
control, sampling actor positions following a general data distribution. Our text-conditioned
model achieves an agent AP of 26.8\%, compared to 5.5\% for TrafficGen, representing a
$\propto$ {}4$\times$ improvement. This confirms that grounding agent placement in a
conditioning prompt provides a strong signal for realistic and accurate vehicle
initialization, substantially outperforming unconditioned placement on the same
maps.
The experiments are carried out on 14688 scenarios from the test split of the Waymo Open Motion dataset \cite{Sun_2020_CVPR_waymo_perception}.

\section{Conclusion}\label{sec:conclusion}
We introduce ScenarioControl, a multimodal conditional method for generating controllable and realistic driving scenarios with text prompts or images. We achieve prompt and visual conditioning through our proposed cross-global attention mechanism for vectorized scenarios that fuse dense multimodal cues with sparse map–agent structures. Additionally, the proposed method supports sensor video generation, producing temporally consistent renderings that visually ground the synthesized scenarios. Through extensive evaluations, we confirm that our control mechanism is favorable for prompt and image conditions, while allowing for diversity and fidelity, effectively bridging structured scenario simulation with realistic sensor-level inputs and human-interpretable guidance.

Our method opens several promising directions for further research. An interesting direction is to leverage controllable scene generation to help autonomous vehicles anticipate traffic situations beyond line of sight information by synthesizing plausible continuations of partially observed environments. Extending our method to full multi-view and long-horizon visual conditioning could further improve consistency. Finally, extending the latent representation to unify initial scene generation and traffic simulation could allow the capture of additional semantics such as intent, weather, or interaction cues -- providing even finer control over scenario structure and agent behavior.

{
    \small
    \bibliographystyle{splncs04}
    \bibliography{main}
}

\end{document}